# R²CNN: Rotational Region CNN for Orientation Robust Scene Text Detection

Yingying Jiang, Xiangyu Zhu, Xiaobing Wang, Shuli Yang, Wei Li, Hua Wang, Pei Fu and Zhenbo Luo

Samsung R&D Institute China - Beijing
{yy.jiang, xiangyu.zhu, x0106.wang, shuli.yang, wei2016.li, hua00.wang, pei.fu, zb.luo}@samsung.com

## Abstract

*In this paper, we propose a novel method called Rotational Region CNN (R²CNN) for detecting arbitrary-oriented texts in natural scene images. The framework is based on Faster R-CNN [1] architecture. First, we use the Region Proposal Network (RPN) to generate axis-aligned bounding boxes that enclose the texts with different orientations. Second, for each axis-aligned text box proposed by RPN, we extract its pooled features with different pooled sizes and the concatenated features are used to simultaneously predict the text/non-text score, axis-aligned box and inclined minimum area box. At last, we use an inclined non-maximum suppression to get the detection results. Our approach achieves competitive results on text detection benchmarks: ICDAR 2015 and ICDAR 2013.*

1. Introduction

   Texts in natural scenes (e.g., street nameplates, store names, good names) play an important role in our daily life. They carry essential information about the environment. After understanding scene texts, they can be used in many areas, such as text-based retrieval, translation, etc. There are usually two key steps to understand scene texts: text detection and text recognition. This paper focuses on scene text detection. Scene text detection is challenging because scene texts have different sizes, width-height aspect ratios, font styles, lighting, perspective distortion, orientation, etc. As the orientation information is useful for scene text recognition and other tasks, scene text detection is different from common object detection tasks that the text orientation should be also be predicted in addition to the axis-aligned bounding box information.

   While most previous text detection methods are designed for detecting horizontal or near-horizontal texts [2,3,4,5,6,7,8,9,10,11,12,13,14], some methods try to address the arbitrary-oriented text detection problem [15,16,17,18,19,20,31,32,33,34]. Recently, arbitrary-oriented scene text detection is a hot research area, which can be seen from the frequent result updates in ICDAR2015 Robust Reading competition in incidental scene text detection [21]. While traditional text detection methods are based on sliding-window or Connected Components (CCs) [2,3,4,6,10,13,17,18,19,20], deep learning based methods have been widely studied recently [7,8,9,12,15,16,31,32,33,34].

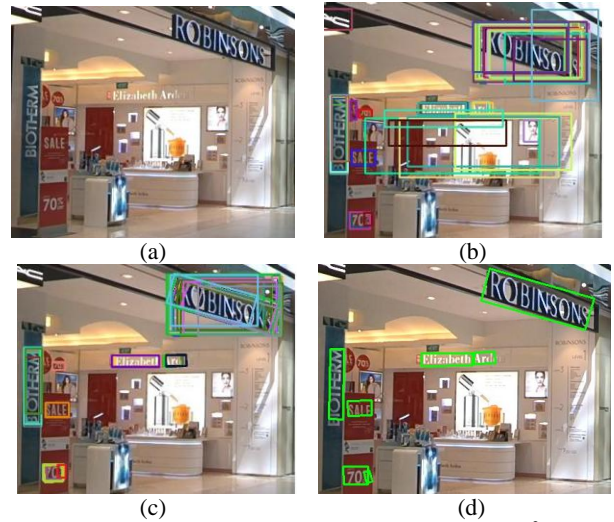

Fig. 1. The procedure of the proposed method R²CNN. (a) Original input image; (b) text regions (axis-aligned bounding boxes) generated by RPN; (c) predicted axis-aligned boxes and inclined minimum area boxes (each inclined box is associated with an axis-aligned box, and the associated box pair is indicated by the same color); (d) detection result after inclined non-maximum suppression.

   This paper presents a Rotational Region CNN (R²CNN) for detecting arbitrary-oriented scene texts. It is based on Faster R-CNN architecture [1]. Figure 1 shows the procedure of the proposed method. Figure 1(a) is the original input image. We first use the RPN to propose axis-aligned bounding boxes that enclose the texts (Figure 1(b)). Then we classify the proposals, refine the axis-aligned boxes and predict the inclined minimum area boxes with pooled features of different pooled sizes (Figure 1(c)). At last, inclined non-maximum suppression is used to post-process the detection candidates to get the final detection results (Figure 1(d)). Our method yields an F-measure of 82.54% on ICDAR 2015 incidental text detection benchmark and 87.73% on ICDAR 2013 focused text detection benchmark.

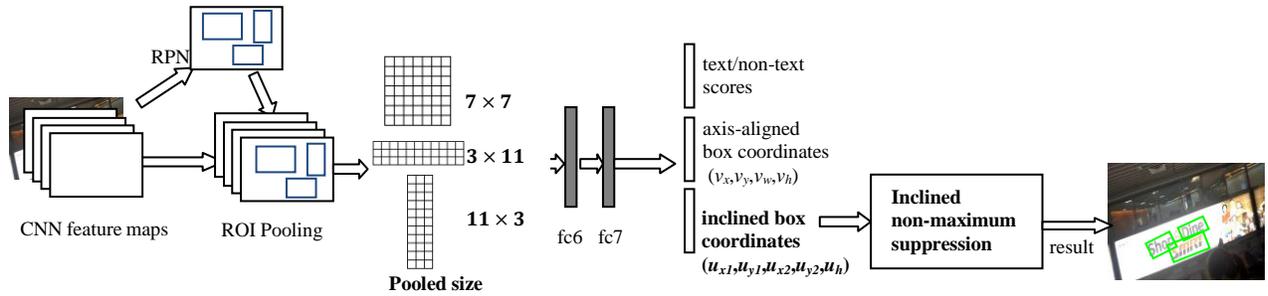

Fig.2. The network architecture of Rotational Region CNN ($R^2$CNN). The RPN is used for proposing axis-aligned bounding boxes that enclose the arbitrary-oriented texts. For each box generated by RPN, three ROIPoolings with different pooled sizes are performed and the pooled features are concatenated for predicting the text scores, axis-aligned box ($v_x,v_y,v_w,v_h$) and inclined minimum area box ($u_{x1},u_{y1},u_{x2},u_{y2},u_h$). Then an inclined non-maximum suppression is conducted on the inclined boxes to get the final result.

The contributions of this paper are as follows:
- We introduce a novel framework for detecting scene texts of arbitrary orientations (Figure 2). It is based on Faster R-CNN [1]. The RPN is used for proposing text regions and the Fast R-CNN model [23] is modified to do text region classification, refinement and inclined box prediction.
- The arbitrary-oriented text detection problem is formulated as a multi-task problem. The core of the approach is predicting text scores, axis-aligned boxes and inclined minimum area boxes for each proposal generated by the RPN.
- To make the most of text characteristics, we do several ROIPoolings with different pooled sizes ($7 \times 7, 11 \times 3, 3 \times 11$) for each RPN proposal. The concatenated features are then used for further detection.
- Our modification of Faster R-CNN also include adding a smaller anchor for detecting small scene texts and using inclined non-maximum suppression to post-process the detection candidates to get the final result.

2. Related Work

The traditional scene text detection methods consist of sliding-window based methods and Connected Components (CCs) based methods [2,3,4,6,10,13,17,18,19,20]. The sliding-window based methods move a multi-scale window through an image densely and then classify the candidates as character or non-character to detect character candidates. The CCs based approaches generate character candidates based on CCs. In particular, the Maximally Stable Extremal Regions (MSER) based methods used to achieve good performances in ICDAR 2015 [21] and ICDAR 2013 [22] competitions. These traditional methods adopt a bottom-up strategy and often needs several steps to detect texts (e.g., character detection, text line construction and text line classification).

The general object detection is a hot research area recently. Deep learning based techniques have advanced object detection substantially. One kind of object detection methods rely on region proposal, such as R-CNN [24], SPPnet [25], Fast R-CNN [23], Faster R-CNN [1], and R-FCN[26]. Another family of object detectors do not rely on region proposal and directly estimate object candidates, such as SSD[27] and YOLO [28]. Our method is based on the Faster R-CNN architecture. In Faster R-CNN, a Region Proposal Network (RPN) is proposed to generate high-quality object proposals directly from the convolutional feature maps. The proposals generated by RPN is then refined and classified with the Fast R-CNN model [23]. As scene texts have orientations and are different from general objects, the general object detection methods cannot be used directly in scene text detection.

Deep learning based scene text detection methods [7,8,9,12,15,16,31,32,33,34] have been studied and achieve better performance than traditional methods. TextBoxes is an end-to-end fast scene text detector with a single deep neural network [8]. DeepText generates word region proposals via Inception-RPN and then scores and refines each word proposal using the text detection network [7]. Fully-Convolutional Regression Network (FCRN) utilizes synthetic images to train the model for scene text detection [12]. However, these methods are designed to generate axis-aligned detection boxes and do not address the text orientation problem. Connectionist Text Proposal Network (CTPN) detects vertical boxes with fixed width, uses BLSTM to catch the sequential information and then links the vertical boxes to get final detection boxes [9]. It is good at detecting horizontal texts but not fit for high inclined texts. The method based on Fully Convolutional Network (FCN) is designed to detect multi-oriented scene texts [16]. Three steps are needed in this method: text block detection by text block FCN, multi-oriented text line candidate generation based on MSER and text line candidates classification. Rotation Region Proposal Network (RRPN) is also proposed to detect arbitrary-oriented scene text [15]. It is based on Faster R-CNN [1]. The RPN is modified to generate inclined proposals with text orientation angle information and the following classification and regression are based on the inclined proposals. SegLink [31] is proposed to detect oriented texts by detecting segments and links. It works well on text lines with arbitrary lengths. EAST [32] is designed to yield fast and accurate text detection in natural scenes. DMPNet [33] is designed to detect text with tighter quadrangle. Deep direct regression [34] is proposed for multi-oriented scene text detection.

Our goal is to detect arbitrary-oriented scene texts. Similar to RRPN [15], our network is also based on Faster R-CNN [1], but we utilize a different strategy other than generating inclined proposals. We think the RPN is qualified to generate text candidates and we predict the orientation information based on the text candidates proposed by RPN.

3. Proposed Approach

In this section, we introduce our approach to detect arbitrary-oriented scene texts. Figure 2 shows the architecture of the proposed Rotational Region CNN ($R^2$CNN). We first present how we formalize the arbitrary-oriented text detection problem and then introduce the details of $R^2$CNN. After that, we describe our training objectives.

3.1. Problem definition

In ICDAR 2015 competition [21], the ground truth of incidental scene text detection is represented by four points in clockwise (x1,y1,x2,y2,x3,y3,x4,y4) as shown in Figure 3(a). The label is at word level. The four points form a quadrangle, which is probably not a rectangle. Although scene texts can be more closely enclosed by irregular quadrangles due to perspective distortion, they can be roughly enclosed by inclined rectangles with orientation (Figure 3(b)). As we think an inclined rectangle is able to cover most of the text area, we approximate the arbitrary-oriented scene text detection task as detecting an inclined minimum area rectangle in our approach. In the rest of the paper, when we mention the bounding box, it refers to a rectangular box.

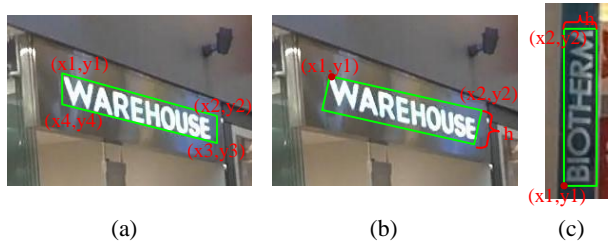

(a)　　　　　　(b)　　　　　　(c)

Fig.3. Detection targets of arbitrary-oriented scene text detection. a) ICDAR 2015 labels the incidental scene texts in the form of four points in clockwise; b) the inclined minimum area rectangle is adopted as the detection targets in our approach; c) another example of the inclined rectangle.

Although the straightforward method to represent a inclined rectangle is using an angle to represent its orientation, we don't adopt this strategy because the angle target is not stable in some special points. For example, a rectangle with rotation angle $90^0$ is very similar to the same rectangle with rotation angle $-90^0$, but their angles are quite different. This makes the network hard to learn to detect vertical texts. Instead of using an angle to represent the orientation information, we use the coordinates of the first two points in clockwise and the height of the bounding box to represent an inclined rectangle (x1,y1,x2,y2,h). We suppose the first point always means the point at the left-top corner of the scene text. Figure 3(b) and Figure 3(c) show two examples. (x1,y1) is the coordinates of the first point (the solid red point), (x2,y2) is the coordinates of the second point in clockwise, and h is the height of the inclined minimum area rectangle.

3.2. Rotational Region CNN ($R^2$CNN)

**Overview.** We adopt the popular two-stage object detection strategy that consists of region proposal and region classification. Rotational Region CNN ($R^2$CNN) is based on Faster R-CNN [1]. Figure 2 shows the architecture of $R^2$CNN. The RPN is first used to generate text region proposals, which are axis-aligned bounding boxes that enclose the arbitrary-oriented texts (Figure 1(b)). And then for each proposal, several ROIPoolings with different pooled sizes ($7 \times 7, 11 \times 3, 3 \times 11$) are performed on the convolutional feature maps and the pooled features are concatenated for further classification and regression. With concatenated features and fully connected layers, we predict text/non-text scores, axis-aligned boxes and inclined minimum area boxes (Figure 1(c)). After that, the inclined boxes are post-processed by inclined non-maximum suppression to get the detection results (Figure 1(d)).

**RPN for proposing axis-aligned boxes.** We use the RPN to generate axis-aligned bounding boxes that enclose the arbitrary-oriented texts. This is reasonable because the text in the axis-aligned box belongs to one of the following situations: a) the text is in the horizontal direction; b) the text is in the vertical direction; c) the text is in the diagonal direction of the axis-aligned box. As shown in Figure 1(b), the RPN is competent for generating text regions in the form of axis-aligned boxes for arbitrary-oriented texts.

Compared to general objects, there are more small scene texts. We support this by utilizing a smaller anchor scale in RPN. While the original anchor scales are (8,16,32) in Faster R-CNN [1], we investigate two strategies: a) changing an anchor scale to a smaller size and using (4,8,16); b) adding a new anchor scale and utilizing (4,8,16,32). Our experiments confirm that the adoption of the smaller anchor is helpful for scene text detection.

We keep other settings of RPN the same as Faster R-CNN [1], including the anchor aspect ratios, the definition of positive samples and negative samples, and etc.

**ROIPoolings of different pooled sizes.** The Faster R-CNN framework does ROIPooling on the feature map with pooled size $7 \times 7$ for each RPN proposal. As the widths of some texts are much larger than their heights, we try to use three ROIPoolings with different sizes to catch more text characteristics. The pooled features are concatenated for further detection. Specifically, we add two pooled sizes: $11 \times 3$ and $3 \times 11$. The pooled size $3 \times 11$ is supposed to catch more horizontal features and help the detection of the horizontal text whose width is much larger than its height. The pooled size $11 \times 3$ is supposed to catch more vertical features and be useful for vertical text detection that the height is much larger than the width.

**Regression for text/non-text scores, axis-aligned boxes, and inclined minimum area boxes.** In our approach, after RPN, we classify the proposal generated by RPN as text or non-text, refine the axis-aligned bounding boxes that contain the arbitrary-oriented texts and predict inclined bounding boxes. Each inclined box is associated with an axis-aligned box (Figure 1(c) and Figure 4(a)). Although our detection targets are the inclined bounding boxes, we think adding additional constraints (axis-aligned bounding box) could improve the performance. And our evaluation also confirm the effectiveness of this idea.

**Inclined non-maximum suppression.** Non-Maximum Suppression (NMS) is extensively used to post-process detection candidates by current object detection methods. As we estimate both the axis-aligned bounding box and the inclined bounding box, we can either do normal NMS on axis-aligned bounding boxes, or do inclined NMS on inclined bounding boxes. In the inclined NMS, the calculation of the traditional Intersection-over-Union (IoU) is modified to be the IoU between two inclined bounding boxes. The IoU calculation method used in [15] is utilized.

Figure 4 illustrates the detection results after two kinds of NMS are performed. Figure 4(a) shows predicted candidate boxes that each axis-aligned bounding box is associated with an inclined bounding box. Figure 4(b) shows the effects of the normal NMS on axis-aligned boxes and Figure 4(c) demonstrates the effects of the inclined NMS on inclined boxes. As show in Figure 4(b), the text in red dashed box is not detected under normal NMS on axis-aligned boxes. Figure 4(d) and Figure 4(e) shows the reason why the inclined NMS is better for inclined scene text detection. We can see that for closely adjacent inclined texts, normal NMS may miss some text as the IoU between axis-aligned boxes can be high (Figure 4(d)), but inclined NMS will not miss the text because the inclined IoU value is low (Figure 4(e)).

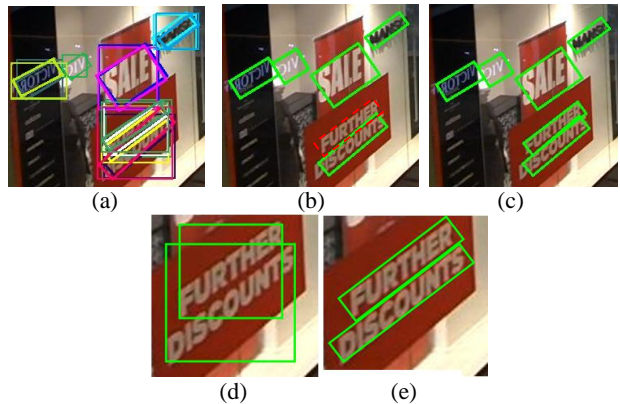

Fig.4. Inclined non-maximum suppression. (a) The candidate axis-aligned boxes and inclined boxes; (b) the detection results based on normal NMS on axis-aligned boxes (the green boxes are the correct detections, and the red dashed box is the box that is not detected); (c) the detection results based on inclined NMS on inclined boxes; (d) an example of two axis-aligned boxes; (e) an example of two inclined boxes.

3.3. Training objective (Multi-task loss)

The training loss on RPN is the same as Faster R-CNN [1]. In this section, we only introduce the loss function of R²CNN on each axis-aligned box proposal generated by RPN.

Our loss function defined on each proposal is the summation of the text/non-text classification loss and the box regression loss. The box regression loss consists of two parts: the loss of axis-aligned boxes that enclose the arbitrary-oriented texts and the loss of inclined minimum area boxes. The multi-task loss function on each proposal is defined as:

$$L(p,t,v,v^*,u,u^*) = L_{\text{cls}}(p,t) \\ + \lambda_1 t \sum_{i\in\{x,y,w,h\}} L_{reg}(v_i, v_i^*) \\ + \lambda_2 t \sum_{i\in\{x1,y1,x2,y2,h\}} L_{reg}(u_i, u_i^*) \quad (1)$$

$\lambda_1$ and $\lambda_2$ are the balancing parameters that control the trade-off between three terms.

The box regression only conducts on text. $t$ is the indicator of the class label. Text is labeled as 1 ($t = 1$), and background is labeled as 0 ($t = 0$). The parameter $p = (p_0, p_1)$ is the probability over text and background classes computed by the softmax function. $L_{\text{cls}}(p,t) = -\log p_t$ is the log loss for true class t.

$v = (v_x, v_y, v_w, v_h)$ is a tuple of true axis-aligned bounding box regression targets including coordinates of the center point and its width and height, and $v^* = (v_x^*, v_y^*, v_w^*, v_h^*)$ is the predicted tuple for the text label. $u = (u_{x1}, u_{y1}, u_{x2}, u_{y2}, u_h)$ is a tuple of true inclined bounding box regression targets including coordinates of first two points of the inclined box and its height, and $u^* = (u^*_{x1}, u^*_{y1}, u^*_{x2}, u^*_{y2}, u^*_h)$ is the predicted tuple for the text label. We use the parameterization for $v$ and $v^*$ given in [24], in which $v$ and $v^*$ specify a scale-invariant translation and log-space height/width shift relative to an object proposal. For inclined bounding boxes, the parameterization of $(u_{x1}, u_{y1})$, $(u_{x2}, u_{y2})$, $(u^*_{x1}, u^*_{y1})$ and $(u^*_{x2}, u^*_{y2})$ is the same with that of $(v_x, v_y)$. And the parameterization of $u_h$ and $u^*_h$ is the same with the parameterization of $v_h$ and $v^*_h$.

Let $(w, w^*)$ indicates $(v_i, v^*_i)$ or $(u_i, u^*_i)$, $L_{reg}(w, w^*)$ is defined as:

$$L_{reg}(w, w^*) = \text{smooth}_{L1}(w - w^*) \quad (2)$$

$$\text{smooth}_{L1}(x) = \begin{cases} 0.5x^2 & \text{if } |x| < 1 \\ |x| - 0.5 & \text{otherwise} \end{cases} \quad (3)$$

4. Experiments

4.1. Implementation details

**Training Data.** Our training dataset includes 1000 incidental scene text images from ICDAR 2015 training dataset [21] and 2000 focused scene text images we collected. The scene texts in images we collected are clear and quite different from the blurry texts in ICDAR 2015. Although our simple experiments show that the additional collected images do not increase the performance on ICDAR2015, we still include them in the training to make our model more robust to different kinds of scene texts. As ICDAR 2015 training dataset contains difficult texts that is hard to detect which are labeled as "###", we only use those readable text for training. Moreover, we only use those scene texts composed of more than one character for training.

To support arbitrary-oriented scene text detection, we augment ICDAR 2015 training dataset and our own data by rotating images. We rotate our image at the following angles (-90, -75, -60, -45, -30, -15, 0, 15, 30, 45, 60, 75, 90). Thus, after data augmentation, our training data consists of 39000 images.

The texts in ICDAR 2015 are labeled at word level with four clockwise point coordinates of quadrangle. As we simplify the problem of incidental text detection as detecting inclined rectangles as introduced in Section 3.1, we generate the ground truth inclined bounding box (rectangular data) from the quadrangle by computing the minimum area rectangle that encloses the quadrangle. We then compute the minimum axis-aligned bounding box that encloses the text as the ground truth axis-aligned box. Similar processing is done to generate ground truth data for images we collected.

**Training.** Our network is initialized by the pre-trained VGG16 model for ImageNet classification [29]. We use the end-to-end training strategy. All the models are trained $20 \times 10^4$ iterations in all. Learning rates start from $10^{-3}$, and are multiplied by $\frac{1}{10}$ after $5 \times 10^4$, $10 \times 10^4$ and $15 \times 10^4$ iterations. Weight decays are 0.0005, and momentums are 0.9. All experiments use single scale training. The image's shortest side is set as 720, while the longest side of an image is set as 1280. We choose this image size because the training and testing images in ICDAR 2015 [21] have the size (width: 1280, height: 720).

4.2. Performance

We evaluate our method on ICDAR 2015 [21] and ICDAR 2013 [22]. The evaluation follows the ICDAR Robust Reading Competition metrics in the form of Precision, Recall and F-measure. The results are obtained by submitting the detection results to the competition website and get the evaluation results online.

*A. ICDAR 2015*

This section introduces our performances on ICDAR 2015 [21]. ICDAR 2015 competition test dataset consists of 500 images containing incidental scene texts with arbitrary orientations. Our method could achieve competitive results of Recall 79.68%, Precision 85.62%, and F-measure 82.54%.

We conduct several experiments to confirm the effectiveness of our design. Table 1 summarizes the results of our models under different settings. We will compare the following models: Faster R-CNN [1], R$^2$CNN-1, R$^2$CNN-2, R$^2$CNN-3, R$^2$CNN-4, and R$^2$CNN-5. We mainly focus on evaluating the influence of the axis-aligned box regression ($\lambda_1$) and the inclined box regression ($\lambda_2$), the influence of anchor scales and NMS strategy, and the impact of different pooled sizes of ROIPoolings. All these models are trained on the same dataset introduced in the last section.

We first perform single-scale testing with all models on ICDAR 2015. The test images keep the original size (width: 1280, height: 720) when performing the testing. We then do multi-scale testing following [30] on R$^2$CNN-3, R$^2$CNN-4, and R$^2$CNN-5. With a trained model, we compute convolutional feature maps on an image pyramid, where the image's short sides are $s \in \{720, 1200\}$. The results of all our designs are better than the baseline Faster R-CNN.

**Axis-aligned box and inclined box.** While Faster RCNN only regresses axis-aligned bounding boxes which is implemented by setting $\lambda_1 = 1$ and $\lambda_2 = 0$ in Equation (1), the detection outputs are axis-aligned boxes. Different from Faster RCNN, R$^2$CNN-1 only regresses inclined boxes ($\lambda_1 = 0$ and $\lambda_2 = 1$ in Equation (1)) and this leads to about 6% performance improvement over Faster R-CNN (F-measure: 62.40% vs. 56.63%). The reason is that the outputs of Faster R-CNN are axis-aligned boxes and the orientation information is ignored. R²CNN-2 regresses both the axis-aligned boxes that enclose the texts and the inclined boxes ($\lambda_1 = 1$ and $\lambda_2 = 1$ in Equation (1)) and leads to another 6% performance improvement over R²CNN-1 (F-measure: 68.49% vs. 62.40%). This means that learning the additional axis-aligned box could help the detection of the inclined box.

Table 1. Results of R² CNN under different settings on ICDAR 2015.

| Approaches | Anchor scales | Axis-aligned box ($\lambda_1$) and inclined box ($\lambda_2$) | Pooled sizes | Inclined NMS | Test scales (short side) | Recall | Precision | F-measure | Time |
|---|---|---|---|---|---|---|---|---|---|
| Faster R-CNN | (8,16,32) | $\lambda_1 = 1, \lambda_2 = 0$ | $7 \times 7$ | | (720) | 59.12% | 54.34% | 56.63% | 0.38s |
| R²CNN-1 | (8,16,32) | $\lambda_1 = 0, \lambda_2 = 1$ | $7 \times 7$ | | (720) | 63.60% | 61.24% | 62.40% | 0.39s |
| R²CNN-2 | (8,16,32) | $\lambda_1 = 1, \lambda_2 = 1$ | $7 \times 7$ | | (720) | 68.22% | 68.75% | 68.49% | 0.4s |
| R²CNN-3 | (4, 8, 16) | $\lambda_1 = 1, \lambda_2 = 1$ | $7 \times 7$ | | (720) | 71.98% | 73.94% | 72.94% | 0.4s |
| | | | | Y | (720) | 72.41% | 76.27% | 74.29% | 0.4s |
| | | | | | (720,1200) | 77.32% | 80.18% | 78.73% | 2.2s |
| | | | | Y | (720,1200) | 78.33% | 83.22% | 80.7% | 2.2s |
| R²CNN-4 | (4, 8,16,32) | $\lambda_1 = 1, \lambda_2 = 1$ | $7 \times 7$ | | (720) | 72.70% | 73.16% | 72.93% | 0.41s |
| | | | | Y | (720) | 72.94% | 75.83% | 74.36% | 0.41s |
| | | | | | (720,1200) | 78.43% | 81.09% | 79.74% | 2.22s |
| | | | | Y | (720,1200) | 79.63% | 84.09% | 81.8% | 2.23s |
| R²CNN-5 | (4, 8,16,32) | $\lambda_1 = 1, \lambda_2 = 1$ | $7 \times 7$, $11 \times 3$, $3 \times 11$ | | (720) | 74.68% | 74.14% | 74.41% | 0.45s |
| | | | | Y | (720) | 74.29% | 76.42% | 75.34% | 0.45s |
| | | | | | (720,1200) | 78.48% | 84.63% | 81.44% | 2.25s |
| | | | | Y | (720,1200) | 79.68 % | 85.62 % | 82.54% | 2.25s |

**Anchor scales.** R²CNN-3 and R²CNN-4 are designed to evaluate the influence of anchor scales on scene text detection. They should regress both the axis-aligned boxes and the inclined boxes ($\lambda_1 = 1$ and $\lambda_2 = 1$ in Equation (1)). R²CNN-3 utilizes smaller anchor scales (4,8,16) compared to the original scales (8,16,32). R²CNN-4 adds a smaller anchor scale to the anchor scales and the anchor scales become (4,8,16,32), which would generate 12 anchors in RPN. Results show that under single-scale test R²CNN-3 and R²CNN-4 have similar performance (F-measure: 72.94% vs. 72.93%), but they are both better than R²CNN-2 (F-measure: 68.49 %). This shows that small anchors could improve the scene text detection performance.

Under multi-scale test, R²CNN-4 is better than R²CNN-3 (F-measure: 79.74% vs. 78.73%). The reason is that scene texts can have more kinds of scales in the image pyramid under multi-scale test and R²CNN-4 with more anchor scales could detect scene texts of various sizes better than R²CNN-3.

**Single pooled size vs. multiple pooled sizes.** R²CNN-5 is supposed to evaluate the effect of multiple pooled sizes. As shown in Table 1, with three pooled sizes ($7 \times 7, 11 \times 3, 3 \times 11$), R²CNN-5 is better than R²CNN-4 with one pooled size ($7 \times 7$) (F-measure: 75.34% vs. 74.36%% under single-scale test and inclined NMS, 82.54% vs. 81.8% under multi-scale test and inclined NMS). This confirm that utilizing more features in R²CNN is helpful for scene text detection.

**Normal NMS on axis-aligned boxes vs. inclined NMS on inclined boxes.** Because we regress both the axis-aligned box and the inclined minimum area box and each axis-aligned box is associated with an inclined box, we compare the performance of normal NMS on axis-aligned boxes and the performance of inclined NMS on inclined boxes. We can see that inclined NMS with R²CNN-3, R²CNN-4 and R²CNN-5 under both single test and multi-scale test consistently perform than their counterparts.

**Test time.** The test times in Table 1 are obtained when doing test on a Tesla K80 GPU. Under single-scale test, our method only increase little detection time compared to the Faster R-CNN baseline.

**Comparisons with state-of-the-art.** Table 2 shows the comparison of R²CNN with state-of-the-art results on ICDAR 2015 [21]. Here, R²CNN refers to R²CNN-5 with inclined NMS. We can see that our method can get competitive results of Recall 79.68%, Precision 85.62% and F-measure 82.54%.

As our approach can be considered as learning the inclined box based on the axis-aligned box, it can be easily adapted to other architectures, such as SSD [27] and YOLO [28].

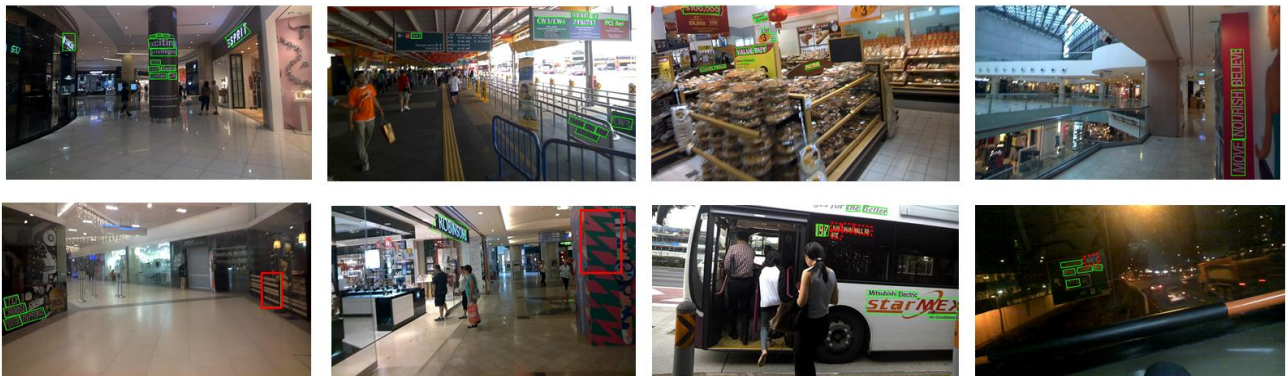

Fig.5. Example detection results of our R²CNN on the ICDAR 2015 benchmark. The green boxes are the correct detection results. The red boxes are false positives. The dashed red boxes are false negatives.

Table 2. Comparison with state-of-the-art on ICDAR2015.

| Approaches | Recall | Precision | F-measure |
|---|---|---|---|
| **R²CNN** | **79.68 %** | **85.62 %** | **82.54 %** |
| Deep direct regression[34] | 80.00% | 82.00% | 81.00% |
| EAST[32] | 78.33% | 83.27% | 80.72% |
| RRPN[15] | 82.17% | 73.23% | 77.44% |
| SegLink[31] | 76.80% | 73.10% | 75.00% |
| DMPNet[33] | 68.22% | 73.23% | 70.64% |
| CTPN[9] | 51.56% | 74.22% | 60.85% |
| MCLAB_FCN[16] | 43.09% | 70.81% | 53.58% |

Figure 5 demonstrates some detection results of our R²CNN on ICDAR 2015. We can see that our method can detect scene texts that have different orientations.

*B. ICDAR 2013*

To evaluate our method's adaptability, we conduct experiments on ICDAR 2013 [22]. ICDAR 2013 test dataset consists of 233 focused scene text images. The texts in the images are horizontal. As we can estimate both the axis-aligned box and the inclined box, we use the axis-aligned box as the output for ICDAR 2013.

We conduct experiments on Faster R-CNN model and R²CNN-5 model trained in last section for ICDAR 2015. Table 3 shows our results and the state-of-the-art results. Our approach could reach the result of F-measure 87.73%. As the training data we used does not include single characters but single characters should be detected in ICDAR 2013, we think our method could achieve even better results when single characters are used for training our model.

To compare our method with the Faster R-CNN baseline, we also do a single-scale test in which the short side of the image is set to 720 pixels. In Table 3, both Faster R-CNN and R²CNN-720 adopt this testing scale. The result is that R²CNN-720 is much better than the Faster R-CNN baseline (F-measure: 83.16 % vs. 78.45%). This means our design is also useful for horizontal text detection.

Table 3. Comparison with state-of-the-art on ICDAR2013.

| Approaches | Recall | Precision | F-measure |
|---|---|---|---|
| CTPN [9] | 83.00% | 93.00% | 88.00% |
| **R²CNN** | **82.59%** | **93.55%** | **87.73%** |
| Deep direct regression[34] | 81.00% | 92.00% | 86.00% |
| SegLink[31] | 83.00% | 87.70% | 85.30% |
| TextBoxes [8] | 82.59% | 87.73% | 85.08% |
| DeepText [7] | 82.79% | 87.17% | 84.93% |
| **R²CNN-720** | **79.73%** | **86.90%** | **83.16%** |
| MCLAB_FCN[16] | 77.81% | 88.14% | 82.65% |
| TextFlow[10] | 75.89% | 85.15% | 80.25% |
| RRPN [15] | 71.89% | 90.22% | 80.02% |
| Faster R-CNN | 74.52% | 82.83% | 78.45% |

Figure 6 shows some detection results on ICDAR 2013. We can see R²CNN could detect horizontal focused scene texts well. The missed text in the figure is a single character.

5. Conclusion

In this paper, we introduce a Rotational Region CNN (R²CNN) for detecting scene texts in arbitrary orientations. The framework is based on Faster R-CNN architecture [1]. The RPN is used to generate axis-aligned region proposals. And then several ROIPoolings with different pooled sizes ($7 \times 7, 11 \times 3, 3 \times 11$) are performed on the proposal and the concatenated pooled features are used for classifying the proposal, estimating both the axis-aligned box and the inclined minimum area box. After that, inclined NMS is performed on the inclined boxes. Evaluation shows that our approach can achieve competitive results on ICDAR2015 and ICDAR2013.

The method can be considered as learning the inclined box based on the axis-aligned box and it can be easily adapted to other general object detection frameworks such as SSD [27] and YOLO [28] to detect object with orientations.

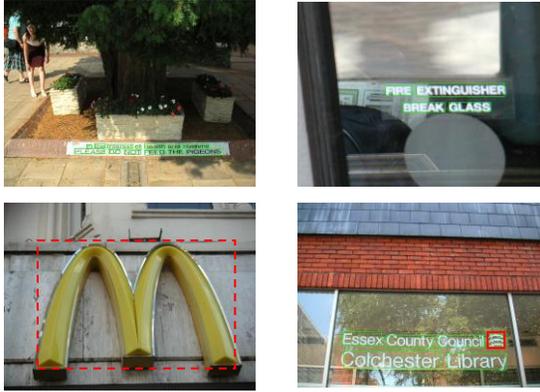

Fig.6. Example detection results of our R²CNN on the ICDAR 2013 benchmark. The green bounding boxes are correct detections. The red boxes are false positives. The red dashed boxes are false negatives.